\documentclass[10pt,twocolumn,letterpaper]{article}

\usepackage{cvpr} 

\usepackage{graphicx}
\usepackage{amsmath}
\usepackage{amssymb}
\usepackage{booktabs}
\usepackage[table,xcdraw]{xcolor}
\usepackage{ dsfont }
\usepackage{multirow}
\usepackage{makecell}
\usepackage{wrapfig}
\usepackage{caption}
\usepackage{array}
\usepackage{soul}

\newcommand{\Y}{\mathcal{Y}}

\newcommand{\I}{\boldsymbol{I}}

\newcommand{\M}{\boldsymbol{M}}
\newcommand{\V}{\mathcal{V}}
\newcommand{\mO}{\mathcal{O}}
\newcommand{\mD}{\mathcal{D}}
\newcommand{\f}{\boldsymbol{f}}

\newcommand{\bb}{\boldsymbol{b}}
\newcommand{\br}{\boldsymbol{r}}

\newcommand{\bv}{\boldsymbol{v}}

\newcommand{\mL}{\mathcal{L}}

\newcommand{\N}{\mathcal{N}}

\newcommand{\argmax}{\operatorname{argmax}}

\newcommand{\BCE}{\operatorname{BCE}}

\newcommand{\e}{\mathrm{e}}

\newtheorem{remark}{Remark}

\newcommand{\myparagraph}[1]{\vspace{-2pt}\medskip\noindent\textbf{#1}}
\newcommand{\point}[0]{\vspace{-0.5pt}\smallskip\noindent$
\bullet$ }

\usepackage[pagebackref,breaklinks,colorlinks]{hyperref}

\usepackage[capitalize]{cleveref}
\crefname{section}{Sec.}{Secs.}
\Crefname{section}{Section}{Sections}
\Crefname{table}{Table}{Tables}
\crefname{table}{Tab.}{Tabs.}

\def\cvprPaperID{1439} 
\def\confName{CVPR}
\def\confYear{2022}

\begin{document}

\title{Open-Vocabulary Instance Segmentation\\via Robust Cross-Modal Pseudo-Labeling}

\author{Dat~Huynh$^{1}$\thanks{This work was done during Dat Huynh's internship at Adobe Research.} \qquad Jason~Kuen$^2$ \qquad Zhe~Lin$^2$ \qquad Jiuxiang~Gu$^2$ \qquad Ehsan~Elhamifar$^1$\\
  \hfill$^1$Northeastern University\hfill
  $^2$Adobe Research\hfill\mbox{ }\\
{\tt\small $^1$\{huynh.dat,e.elhamifar\}@northeastern.edu \qquad  $^2$\{kuen, zlin, jigu\}@adobe.com}}

\maketitle

\begin{abstract}
Open-vocabulary instance segmentation aims at segmenting novel classes without mask annotations.
It is an important step toward reducing laborious human supervision.
Most existing works first pretrain a model on captioned images covering many novel classes and then finetune it on limited base classes with mask annotations.
However, the high-level textual information learned from caption pretraining alone cannot effectively encode the details required for pixel-wise segmentation.
To address this, we propose a cross-modal pseudo-labeling framework, which generates training pseudo masks by aligning word semantics in captions with visual features of object masks in images.
Thus, our framework is capable of labeling novel classes in captions via their word semantics to self-train a student model.
To account for noises in pseudo masks, we design a robust student model that selectively distills mask knowledge by estimating the mask noise levels, hence mitigating the adverse impact of noisy pseudo masks.
By extensive experiments, we show the effectiveness of our framework, where we significantly improve mAP score by 4.5\% on MS-COCO and 5.1\% on the large-scale Open Images \& Conceptual Captions datasets compared to the state-of-the-art.\footnote{Code is available at \url{https://github.com/hbdat/cvpr22_cross_modal_pseudo_labeling}.}
\end{abstract}

\section{Introduction}
Instance segmentation is a crucial yet challenging task of segmenting all objects in an image with applications in autonomous driving, surveillance systems, and medical imaging.
Segmentation works have achieved impressive results thanks to advances in training high capacity models with large amounts of mask annotations \cite{Cordts:CVPR16,OpenImages:16,Gupta:CVPR19,Zamir:CVPRW19}.
To be specific, most methods adopt a two-stage object detection architecture \cite{Ren:CVPR15} for this task by learning an additional mask head to segment objects within box proposals \cite{He:TPAMI20,Liu:CVPR18,Chen:CVPR19,Huang:CVPR19}.
Recent works focus on high-quality mask segmentation by increasing the prediction resolutions using dynamic networks \cite{Arnab:CVPR17,Tian:ECCV20} or boundary refinement \cite{Kirillov:CVPR20,Zhang:CVPR21,Tang:CVPR21}.
Despite their success, these works all require costly mask annotations of every class.
As a result, it is difficult to scale such systems to hundreds or thousands of classes due to their high mask annotation costs for training.
In this work, we aim to significantly reduce the amount of mask supervision by segmenting novel classes using low-cost captioned images.

\begin{figure}[t]
\centering
\includegraphics[width=0.99\linewidth]{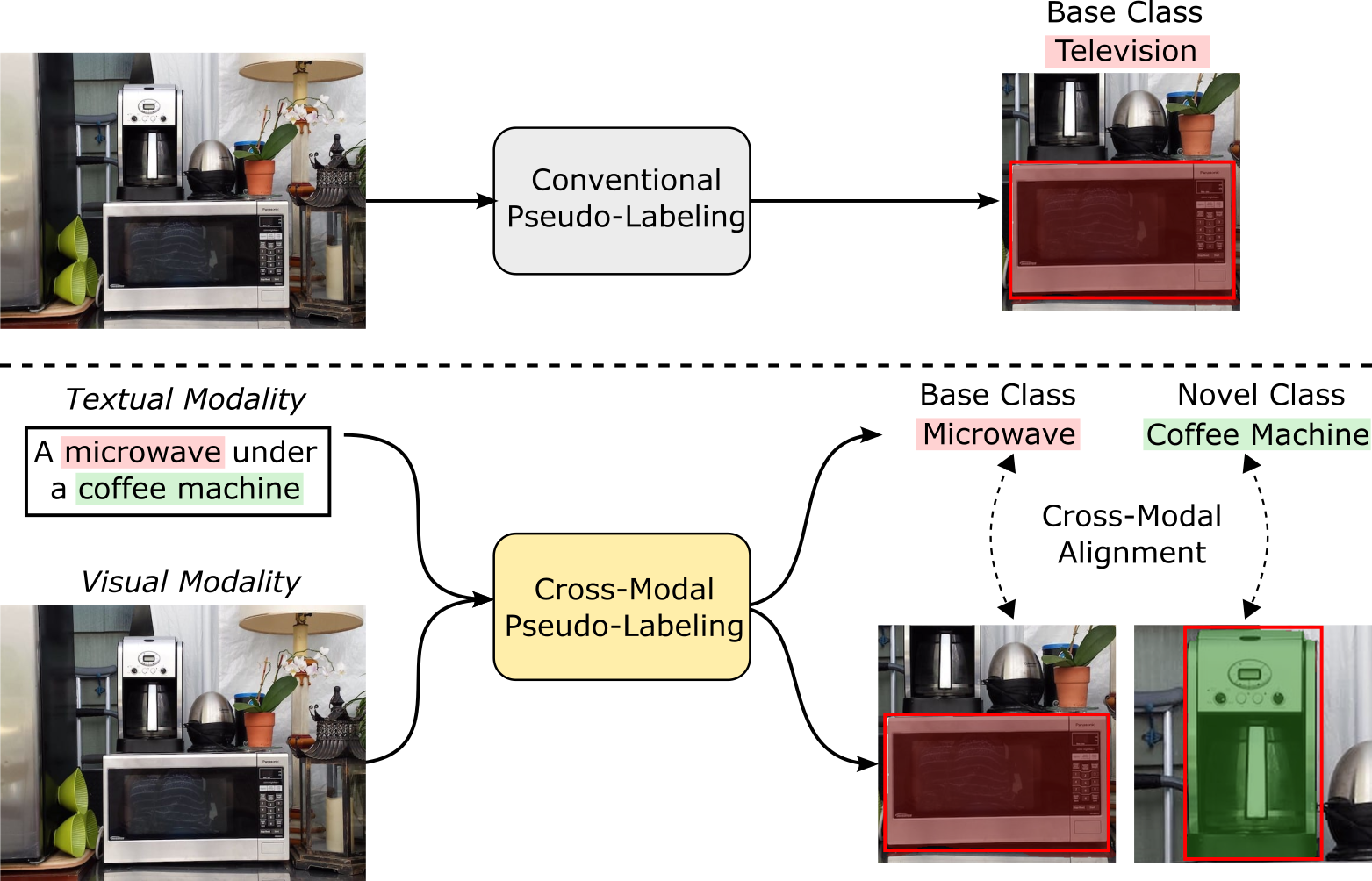}
\caption{
\small{
Conventional pseudo-labeling (\textbf{top}) only segments objects based on visual modality, which produces incorrect labels and misses novel object classes.
Our method (\textbf{bottom}) leverages both visual and textual modalities by aligning semantics of caption words with visual features of object masks to correctly label objects and generalize to novel classes without mask annotations.
}
}
\label{fig:overview}
\vspace{-5mm}
\end{figure}

One of the most popular ways to increase the number of segmentation classes is partially-supervised learning.
It utilizes weak image-level \cite{Hsu:NeurIPS19,Ahn:CVPR19,Arun:ECCV20} or box-level \cite{Hu:CVPR18,Biertimpel:ICCV20,Zhou:CVPR20,Tian:CVPR21,Lee:CVPR21,Wang:CVPR21} supervision to segment objects that have no mask annotations, thus lowering the annotation costs.
Despite the successes of partially-supervised methods, they can only segment the classes covered by the image/box-level annotation and not a wide general range of novel classes.

Different from previous approaches that are limited to classes with mask annotations, zero-shot instance/semantic segmentation aims to segment novel classes without training samples via high-level semantic descriptions such as word embeddings.
However, current zero-shot approaches on both object detection \cite{Bansal:ECCV18,Rahman:AAAI20,Zhu:CVPR20} and instance segmentation \cite{Zheng:CVPR21} suffer from low novel-class performances as high-level word embeddings cannot effectively encode fine-grained shape information.
To overcome this, the recent \texttt{OVR}~\cite{Zareian:CVPR21} work pretrains a visual backbone on captioned images to learn rich visual features.
As the backbone of \texttt{OVR} encodes the visual appearances of many novel classes in captions, finetuning it on the detection task significantly improves the performance of novel classes.
Despite its effectiveness for detection, we argue that backbone pretraining has limited effects on instance segmentation since mask predictions are ignored and not learned during caption pretraining.


In this paper, we address \textit{instance segmentation of novel classes unknown during training} by directly self-training our model to segment objects in captioned images without any mask annotations.
We introduce a robust cross-modal pseudo-labeling framework that aligns textual and visual modalities in captioned images to create caption-driven pseudo masks and generalize to novel classes beyond base classes.
Specifically, we train a teacher model on base classes and use this model to select object regions whose visual features are most compatible with the semantics of words in captions.
The regions are further segmented into pseudo masks for object words in captions.
We then distill pseudo masks into a robust student, which jointly learns segmentation and estimates pseudo-mask noise levels to downweight incorrect teacher predictions.
Finally, we evaluate our segmentation performances on MS-COCO and Open Images \& Conceptual Captions datasets.
We qualitatively demonstrate our generalization ability on truly novel classes, which never appear in most segmentation datasets.

The contributions of this paper are as follows:

\point
We propose a novel cross-modal pseudo-labeling framework to generate caption-driven pseudo masks and fully utilize captioned images for segmentation training without requiring instance mask annotations.

\point
Our method is designed to work with novel classes by selecting regions whose visual features are most compatible with the semantics of novel classes and segmenting these regions into pseudo masks to self-train a student model.

\point
We explicitly capture the reliability of pseudo masks via our robust student model. For pseudo masks with high mask noises, we downweight the loss to avoid error propagation when objects cannot be grounded in images.

\point
To show the effectiveness of our method, we conduct extensive experiments on MS-COCO and the large-scale Open Images \& Conceptual Captions datasets.

\section{Related Works}
\myparagraph{Partially Supervised Learning.}
Due to the high cost of mask annotations \cite{Bearman:ECCV16}, learning segmentation with weak supervision has attracted strong interest recently.
Given bounding box annotations, \cite{Khoreva:CVPR17,Hsu:NeurIPS19,Zhou:CVPR20,Tian:CVPR21,Lan:ICCV21} exploit pixel-wise similarity to infer object masks while \cite{Hu:CVPR18,Kuo:ICCV19,Biertimpel:ICCV20,Fan:ECCV20} learn to share mask knowledge between mask and box supervision to enhance performances.
Whereas, \cite{Zhou:CVPR18,Ahn:CVPR19,Ge:ICCV19,Zhu:CVPR19,Cholakkal:CVPR19,Arun:ECCV20,Shen:ICCV21} leverage image-level labels by analyzing classification scores in image regions to estimate object masks.
Recently, \cite{Laradji:ICIP20,Cheng:Arxiv21,Li:Arxiv21} have explored point-wise supervision and learn from only a few background/foreground pixel annotations.
Unlabeled images can also be used to improve performances by considering confident predictions as annotations of these images for training \cite{Radosavovic:CVPR18,Wang:ECCV18,Sohn:Arxiv20,Li:ECCV20,Zoph:NeurIPS20,Xu:ICCV21,Liu:ICLR21}.
However, these works assume certain forms of weak annotations are available for all classes, thus cannot generalize to a wide range of novel classes that may have no annotations at all.

\myparagraph{Zero-Shot Learning}. To generalize toward novel classes without any training annotations, most zero-shot works \cite{Xian:CVPR17,Schonfeld:CVPR19,Felix:ECCV18,Jiang:ICCV19,Atzmon:CVPR18,Gong:CVPR18,Huynh-multiatt:CVPR20,Huynh-finegrained:CVPR20,Huynh:NeurIPS20,Huynh:ICCV21} focus on image recognition. Recent works have explored zero-shot object detection by learning to distinguish between background and novel object regions \cite{Bansal:ECCV18,Rahman:AAAI20}, synthesizing unseen class features \cite{Zhu:CVPR20} or using richer textual descriptions \cite{Li:AAAI19}.
For pixel-level mask prediction, \cite{Xian:CVPR19,Kato:ICCVW19,Hu:NeurIPS20,Tian:ICM20,Li:NeurIPS20,Zhang_prototype:ICCV21,Baek:ICCV21,Cheng:ICCV21} perform zero-shot semantic segmentation while \cite{Zheng:CVPR21} tackles the challenging zero-shot instance segmentation task.
Since these zero-shot methods only have access to base class annotations, they perform poorly on novel classes.
Although \cite{Bucher:NeurIPS19,Rahman:ICCV19,Pastore:CVPRW21} apply self-training on unlabeled data from novel classes to improve performances, they only address semantic segmentation and cannot distinguish different object instances in an image.
Moreover, they make a strong assumption that unlabeled samples always belong to a restricted set of classes known during training.

\myparagraph{Vision-Language Pretraining}, on the other hand, aims to learn from captioned images containing a wide range of classes.
Most works focus on learning visual backbones that encode rich visual information from caption-image pairs and finetuning
them on downstream tasks.
Specifically, \cite{Tan:EMNLP19,Lu:NeurIPS19,Li:ACL20,Li:AAAI20,Chen:ECCV20} employ pretrained language models and object detectors to learn visual features well aligned with the embeddings of caption words.
Recent works \cite{Yuan:CVPR21,Desai:CVPR21} improve training efficiency by removing the need for object detectors and scale to hundreds of millions of samples for substantial performance gains \cite{Radford:ICML21}.
Moreover, \cite{Zareian:CVPR21} proposes a novel open-vocabulary learning task and shows that pretrained visual features improve not only the detection performance on base classes but also novel classes.
However, backbone pretraining alone cannot exploit captioned images for segmentation, as the model is not trained explicitly to segment the objects in captioned images.

\myparagraph{Learning with Noisy Annotations}.
Although learning with noisy training samples collected from the web or annotated by machine can also significantly reduce the annotation cost, \cite{Zhang:ICLR17} shows that deep neural networks can easily fit random label noises.
Thus, most works address this by regulating the loss function \cite{Natarajan:NIPS13,Sanchez:ICML19,Wang:ICCV19,Zhou:ICCV21,Zhang_dual_graph:CVPR21,Xu:CVPR21,Ortego:CVPR21,Collier:CVPR21,Zhu:CVPR21}, denoising training samples \cite{Veit:CVPR17,Yi:CVPR19,Li:ICCV21}, or utilize additional unlabeled data for regularization \cite{Ding:WACV18,Li:ICLR20}.
As these methods are not applicable to the segmentation task, \cite{Kendall:NIPS17,Hu:NeurIPS20} propose to capture uncertainty in mask predictions to regulate pixel-wise segmentation loss, thus reducing the impacts of noisy annotations.
However, they can only estimate noise from the mask annotations belonging to base classes thus are ineffective for novel classes without mask annotations.

\section{Robust Cross-Modal Pseudo-Labeling\\on Captioned Images}
This section describes our robust cross-modal pseudo-labeling framework, which utilizes caption-image pairs to produce pseudo masks and self-trains a student model. We first describe the problem setting and then present different components in our framework.

\begin{figure*}[t]
\centering
\includegraphics[width=0.8\linewidth]{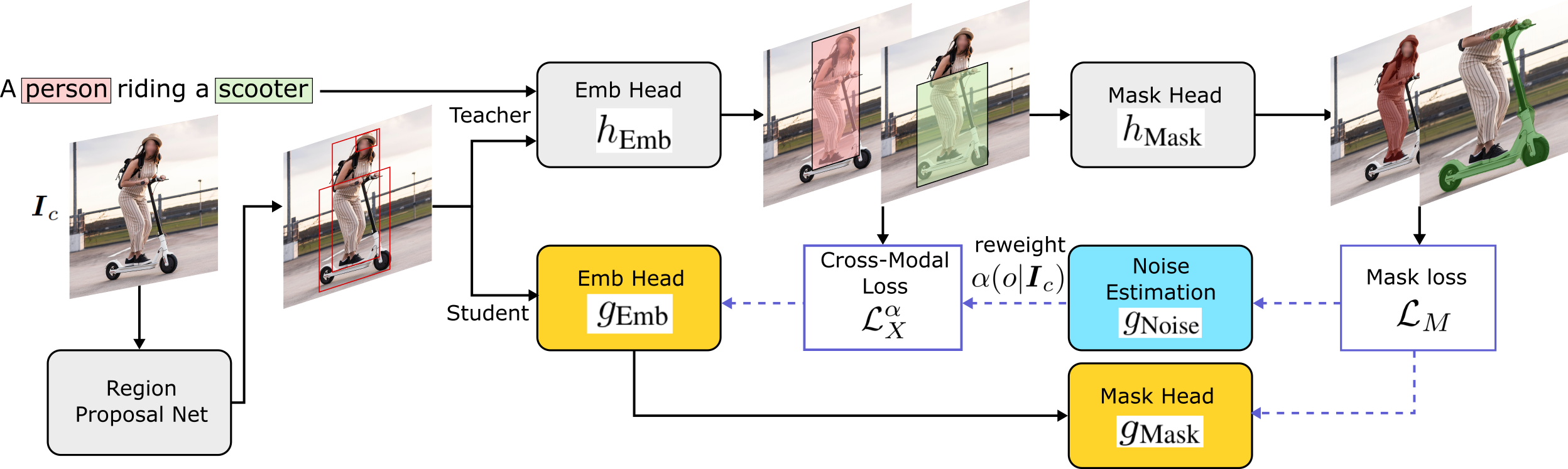}
\vspace{-2mm}
\caption{
\footnotesize{
Given an image $\I_c$ and the set of objects in captions $\mO_c$, we first generate region proposals. We then find the regions that maximize the scores of the teacher embedding head ($h_{\text{Emb}}$) for each object in the caption.
We further segment objects within these regions into pseudo masks using the teacher's mask head ($h_{\text{Mask}}$). Finally, the student embedding ($g_{\text{Emb}}$) and mask ($g_{\text{Mask}}$) heads are trained via cross-modal and mask losses, respectively. The cross-modal loss is also
reweighted based on the pseudo-mask noise levels learned from our pseudo-mask loss.
}
}
\label{fig:schematic}
\vspace{-4mm}
\end{figure*}

\subsection{Problem Setting}
Let $\mathcal{D}_B=\{(\I_m, \Y_m)\}_{m=1}^{N_B}$ be the set of training images and instance annotations for a limited set of base classes $\mathcal{V}_B$.
Each image $\I_m$ is associated with a set of ground-truth (GT) annotations $\Y_m$, which comprises instance masks and their corresponding object classes.
In order to segment novel classes, we leverage additional images $\mathcal{D}_C=\{(\I_c, \Y_c)\}_{c=1}^{N_C} $ with only image-level captions.
Each image $I_c$ is annotated with a caption from which we can extract a set of object nouns $\mathcal{O}_c \subset \Y_c$ in each caption.
Since caption annotations are relatively inexpensive to source, the set of caption classes, $|\mathcal{V}_C|$, is significantly larger than base classes, $|\mathcal{V}_C| \gg |\mathcal{V}_B|$, which is the key ingredient to improve the segmentation of novel classes.

We follow \cite{Zareian:CVPR21} to construct a set of target classes, $\mathcal{V}_T$, without any mask annotations and unknown to the model during training.
These classes are merely used as a proxy to evaluate the segmentation performance of novel classes during test time.
Our model can recognize a much larger number of novel classes, by using the high-level semantic embeddings $\{\bv_o\}$, for all object classes $o \in \mathcal{V}_B\cup\mathcal{V}_C\cup\mathcal{V}_T$, from a pretrained \texttt{BERT} model \cite{Devlin:NAACL19}.
Given the BERT embeddings, we transfer the knowledge from base/caption to target classes via class semantic similarity.

\subsection{Proposed Method}
In this section, we present our proposed cross-modal pseudo-labeling framework for open-vocabulary instance segmentation.
For each caption-image pair, we generate pseudo masks by selecting the mask predictions whose visual features are most compatible with semantic embeddings of object words in captions.
We first construct a teacher model with an embedding head for classification and a class-agnostic mask head for segmentation.
Then, we distill the mask knowledge from teacher predictions and captions into a robust student model which jointly learns from pseudo masks and estimates mask noise levels to downweight unreliable pseudo masks.

\subsubsection{Designing Teacher Model}
To effectively extract mask supervision from captioned images, we first introduce a teacher model, $h$, capable of segmenting novel classes based on the word embeddings of these classes.
Following \cite{Zareian:CVPR21}, we build upon a two-stage detection framework, \texttt{Mask R-CNN} \cite{He:TPAMI20}.
To be specific, we train a class-agnostic region proposal network, $p$, to select a set of region proposals in each image:
$\{\br_i\}_{i=1}^{N_R} = p(\I)$.

Given the region proposals, our goal is to classify them to any classes mentioned in the captions extending beyond base classes.
Therefore, we replace the conventional fully-connected layer in the classification head of \texttt{Mask R-CNN} with an embedding head $h_{\text{Emb}}$.
Here, $h_{\text{Emb}}$ maps the region features into the semantic space of word embeddings.
With the embedding head, the score of class $o$ for each region is computed as inner-product between the word embedding of the class and the region's visual feature:
\begin{equation}
\bv^\top_oh_{\text{Emb}}(\f^{\I}_{\br}) \quad \forall \br \in p(\I),
\end{equation}
where $\bv_o$ is the word embedding for class $o$, $\f^{\I}_{\br}$ is the visual feature of region $\br$ extracted from the visual backbone using \texttt{RoIAlign} \cite{He:TPAMI20} and
$h_{\text{Emb}}(\f^{\I}_{\br})$ is the visual embedding of the region.
To simplify the notation, we drop the super-script $\I$ in $\f^{\I}_{\br}$ which can be inferred from the context.
By learning a joint embedding space between visual features and the word embeddings, the teacher can generalize to novel classes without training samples by measuring the compatibility between visual and textual features.
We also define the background embedding to be a fixed zero vector, which has been shown to outperform other variants \cite{Zareian:CVPR21}.
Thus, a region proposal is considered background if its class scores are lower than the background score. 
In addition, we also learn a class-agnostic Mask R-CNN-based head to segment object in each region as,
$h_{\text{Mask}}(\f_{\br})$,
where
$h_{\text{Mask}}(\cdot)$
is a mask head predicting mask logit scores.
To train both embedding and mask heads of the teacher, we adopt the ground-truth loss, $\mL_{GT}$, consisting of standard detection and segmentation losses as in \cite{He:TPAMI20}.

Although the teacher can segment novel classes, it cannot effectively perform this and often miss-classifies novel classes due to their lack of training annotations.
To provide additional supervision for novel classes without incurring high annotation costs, we propose a cross-modal pseudo-learning method that uses the semantic information of caption words to guide teacher predictions and generates pseudo masks for self-training a student model.

\subsubsection{Cross-Modal Pseudo-Labeling}
To boost the teacher's performance in novel classes, we combine the teacher model with caption guidance and explicitly constrain teacher predictions on what objects and where to construct the pseudo masks for training a student model, $g$.
We first leverage captions to identify objects in images.
For simplicity, we extract object nouns in each caption, $\mathcal{O}_c \subset \Y_c$, as words that are descendants of `Object' node in the WordNet hierarchy, which is inspired by \cite{Gupta:CVPR19}.
To localize these object words in images, we propose a \textit{cross-modal alignment} step that selects the regions whose features are most compatible with the word embeddings of object nouns in captions as following:
\begin{equation}
\bb_o = \underset{\br \in p(\I_c)}{\argmax}\big(\bv^\top_oh_{\text{Emb}}(\f_{\br})\big) \quad \forall o \in \mO_c,
\end{equation}
where the $\bb_o$ is the \textit{aligned object region} for object $o$ w.r.t. its word embedding $\bv_o$ and visual embedding from the teacher, $h_{\text{Emb}}(\f_{\br})$.
As our pseudo labeling procedure is guided by the word semantics in captions, we specifically search for objects in captions and generalize to novel classes based on their word embeddings.
Following recent works on weakly-supervised learning \cite{Tang:CVPR17,Ye:ICCV19}, we select the highest confident bounding box for each object to minimize false-positive predictions.

Given the set of aligned object regions, we introduce a cross-modal loss, $\mL_{X}$, which trains the student to identify these regions as their positively-matched caption words:
\begin{align}
\mL_{X}(\Y_c|\I_c;g) = -\sum_{\substack{o\in \mO_c}} \log \frac{\e^{\bv^\top_o g_{\text{Emb}}(\f_{\bb_o})}}{\sum\limits_{w \in \V_C} \e^{\bv^\top_{w} g_{\text{Emb}}(\f_{\bb_o})}},
\label{eq:cross_modal_loss}
\end{align}
where $g_{\text{Emb}}$ is the student embedding head.
For each aligned object region $\bb_o$, the student maximizes its scores of object words in captions and minimizes the scores of other irrelevant words $w$ via Softmax normalization.
The information from both word embeddings $\{\bv_o\}_{o\in \mO_c}$ (textual modality) and aligned object regions $\{\f_{\bb_o}\}_{o\in\mO_c}$ (visual modality) is distilled into the student embedding head to expand the student's knowledge about the novel classes in captions.

Cross-modal loss works by acting on the student embedding head, but it disregards the mask head that is critical for segmentation.
Next, we propose to obtain the pseudo masks from the teacher and estimate the noise levels of such masks.
Our method provides supervision for the student mask head, in addition to regulating the cross-modal loss.
\vspace{-2.0mm}

\subsubsection{Estimating Pseudo-Mask Noises}
\begin{figure}[t]
\centering
\includegraphics[width=0.8\linewidth]{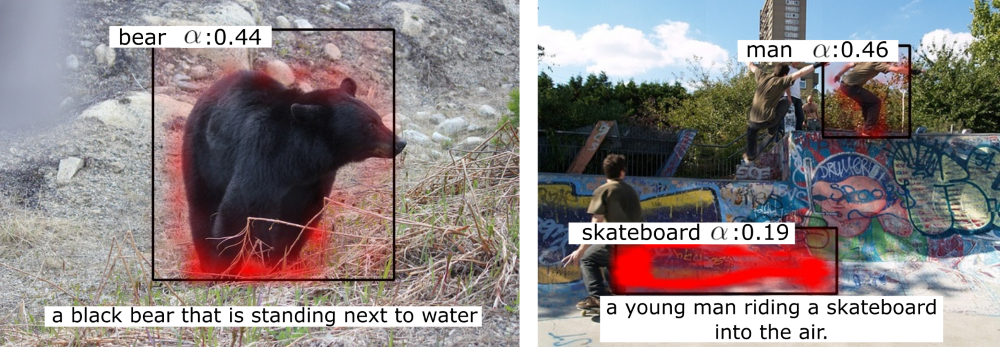}
\caption{
\footnotesize{
Visualization of pseudo mask noise levels and their reliability scores for the objects mentioned in captions.
}
}
\label{fig:mask_noise}
\vspace{-5mm}
\end{figure}

Given aligned object regions, we turn them into pseudo masks by applying the teacher mask head on these regions:
\begin{equation}
\M_{o} = \mathds{1}_{\geq 0}[h_{\text{Mask}}(\f_{\bb_o})] \quad \forall o \in \mO_c,
\end{equation}
where $\mathds{1}_{\geq 0}[\cdot]$ is an indicator function which outputs 1 if a pixel prediction is positive and 0 otherwise to binarize mask predictions.
Naively, we can train the student model to mimic the exact pseudo masks at each pixel as:
\begin{equation}
\sum_{o \in \mO_c}\sum_{x,y} \mathcal{L}_{\text{BCE}}\big(\M^{xy}_o|g_{\text{Mask}}^{xy}(\f_{\bb_o})\big),
\end{equation}
where $\BCE$ is the binary cross-entropy loss for pixel logit predictions, $\M^{xy}_o$ is the pseudo masks at pixel $(x,y)$ and $g_{\text{Mask}}^{xy}$ is the student mask predictions at the pixel.
However, not all objects in captions can be correctly detected
/segmented due to the errors in teacher predictions, as shown in Figure \ref{fig:mask_noise}.
Thus, minimizing this pixel-wise loss propagates the errors from pseudo masks to the student mask head and degrades its performance.
To account for errors in pseudo labels, we propose to estimate the noise level in pseudo masks.
Specifically, the student predicts an additional noise value for each pixel in pseudo masks following \cite{Kendall:NIPS17,Hu:NeurIPS20}.
We assume that each pixel in a pseudo mask is corrupted by a Gaussian noise whose variances can be estimated via the visual features of the aligned object region.
Thus, we can learn to estimate the pixel-wise noise as:
\begin{equation}
\begin{gathered}
\mL_{M}(\Y_c|\I_c,g)= \sum_{o \in \mO_c}\sum_{x,y} \mathcal{L}_{\text{BCE}}\big(\M^{xy}_o|g_{\text{Mask}}^{xy}(\f_{\bb_o})+\epsilon^{xy}_{o}\big)\\
\epsilon^{xy}_{o} \sim \N\big(0,g_{\text{Noise}}^{xy}(\f_{\bb_o})\big),
\end{gathered}
\label{eq:mask_noise}
\end{equation}
where $g_{\text{Noise}}$ is a neural network predicting the noise levels from the visual features of aligned object regions $\f_{\bb_o}$, and $\epsilon^{xy}_{o}$ is the noise value for the pixel $(x,y)$ of object $o$ sampled from the Gaussian distribution, $\N$, parameterized by $g_{\text{Noise}}$.
Pseudo masks with segmentation errors, which are difficult to learn by the student, would drive $g_{\text{noise}}$ to estimate high noise levels to fit these errors.
As such, our framework not only trains the student mask head on pseudo masks but also estimates pseudo-mask noise to regulate the training loss and account for possible segmentation errors of the teacher.

With the ability to estimate pseudo-mask noises, we utilize this to improve cross-modal loss in the next section.

\subsubsection{Training Robust Student Model}
Since both the student and teacher models are unaware of the correct novel object masks due to the lack of annotations, we propose to consider mask noises as a proxy on how reliable the pseudo masks are.
We compute the noise level of each pseudo mask as the average of pixel noise: $\sum_{x,y}g_{\text{Noise}}^{xy}(\f_{\bb_o})/|\bb_o|$ where $|\bb_o|$ is the number of pixels in region $\bb_o$.
Then we assign a \textit{reliability score}, $\alpha(o|\I_c)$, for each object in captions as the inverse of its average noise level, to indicate the mask reliability:
\begin{equation}
\alpha(o|\I_c) = \frac{\eta}{\sum_{x,y}g_{\text{Noise}}^{xy}(\f_{\bb_o})/|\bb_o|} \quad \forall o\in \mO_c,
\end{equation}
where $\eta$ is a constant value set to the smallest average noise level across all captioned images\footnote{We determine $\eta$ by training our method on a subset of images and set the smallest average noise level during training to be $\eta$.}.
With $\eta$ as the reference, we assign low weights to high-noise predictions while upweighting the clean pseudo masks with low noise levels.

\myparagraph{Objective Function}. Finally, we train a robust student model on datasets of caption and base classes as:
\begin{equation}
\begin{aligned}
\min_{g = \{g_{\text{Emb}},g_{\text{Mask}},g_{\text{Noise}}\}} \sum_{c \in \mD_C}\Big[\mL_{M}(\Y_c|\I_c;g) + \mL^{\alpha}_{X}(\Y_c|\I_c;g)\Big]\\
+\sum_{m \in \mD_B}\mL_{GT}(\Y_m|\I_m;g),
\end{aligned}
\end{equation}
where $\mL^{\alpha}_{X}$ is the cross-modal loss in Eq. \eqref{eq:cross_modal_loss} modified to reweight
its term as: $\alpha(o|\I_c) \times \log \frac{\e^{\bv^\top_o g_{\text{Emb}}(\f_{\bb_o})}}{\sum_{w \in \V_C} \e^{\bv^\top_{w} g_{\text{Emb}}(\f_{\bb_o})}}$ for each object, $o\in \mO_c$ .
Thus, we effectively downweight the cross-modal loss on noisy predictions to avoid the error propagation from teacher to student.

\begin{remark}
As the student is trained with cross-modal pseudo-labeling that leverages novel-class information from captioned images, it is able to surpass the teacher's performance.
This is different from conventional knowledge distillation works, where the student is bounded by the teacher's performance.
\end{remark}

\vspace{-3.5mm}

\section{Experiments}
We evaluate our proposed method, which is referred to as \texttt{XPM} for Cross(X)-modal Pseudo Mask, for object detection and instance segmentation on MS-COCO and Open Images \& Conceptual Captions datasets.
Below, we discuss dataset statistics, evaluation metrics, baselines, and implementation details.
We then present and analyze our performances on both base and target classes under various settings.
Finally, we demonstrate the importance of each proposed component via ablation study and show how our noise estimation approach compares with other variants.

\begin{table*}[t]
\centering
\small
\caption{
\footnotesize{
Object Detection (mAP) performances trained with bounding-box or mask supervision on base classes in MS-COCO under constrained setting, which outputs either base or target classes, and generalized setting, which must predict all classes. Improvements w.r.t. to other baselines are highlighted in \textbf{\color{blue}blue}.
* indicates performances reported in \cite{Zareian:CVPR21} while we implement others.
}
}
\vspace{-4pt}
\resizebox{0.95\textwidth}{!}{
\setlength{\tabcolsep}{2pt}{
\begin{tabular}{c|c c|c c c|c c|c c c}
\hline
\multirow{3}{*}{Method} & \multicolumn{5}{c|}{Bounding Box Supervision} & \multicolumn{5}{c}{Instance Mask Supervision}\\
\cline{2-11}
& \multicolumn{2}{c|}{Constrained} & \multicolumn{3}{c|}{Generalized} & \multicolumn{2}{c|}{Constrained} & \multicolumn{3}{c}{Generalized} \\
\cline{2-11}
& \;\; Base \;\; & \;\;\; Target \;\; & \;\; Base \;\; & \;\; Target \;\; & \;\; All \;\; & \;\; Base \;\; & \;\; Target \;\; & \;\; Base \;\; & \;\; Target \;\; & \;\; All \;\; \\
\hline
\multicolumn{11}{c}{\textit{Zero-Shot Training}} \\
\hline
$\texttt{SB}^{*}$ \cite{Bansal:ECCV18}& 29.7&	0.7&	29.2&	0.3&	24.9 & - &	- &	- &	- &	- \\
$\texttt{BA-RPN}^{*}$ \cite{Zheng:CVPR21}& - &	11.4 &	46.5&	4.8&	35.6 & -&	-&	-&	-&	- \\
\hline
\multicolumn{11}{c}{\textit{Caption Pretraining with \cite{Zareian:CVPR21}}} \\
\hline
\texttt{OVR} \cite{Zareian:CVPR21}& 46.8 &	27.5 &	46.0 &	22.8 &	39.9 & 47.2&	25.9&	\textbf{46.7}&	20.7&	39.9 \\
\texttt{SB} \cite{Bansal:ECCV18}& 46.9&	26.9&	46.3&	21.2&	39.7 & 45.9&	25.7&	45.3&	19.6&	38.6 \\
\texttt{BA-RPN} \cite{Zheng:CVPR21}& 46.8&	26.0&	46.2&	20.7&	39.5 & 46.0&	25.0&	45.5&	19.3&	38.7 \\
\texttt{OVR+OMP} \cite{Biertimpel:ICCV20}& -&	-&	-&	-&	- & 34.1&	16.9&	33.2&	10.0&	27.1\\
\hline
\multicolumn{11}{c}{\textit{Pseudo-Labeling}} \\
\hline
\texttt{Soft-Teacher} \cite{Xu:ICCV21}& 47.4&	18.8&	47.1&	12.4&	38.0 & 46.6&	16.0&	46.2&	10.4&	36.8 \\
\texttt{Unbiased-Teacher} \cite{Liu:ICLR21}& \textbf{47.5}&	20.5&	\textbf{47.2}&	13.8&	38.4 & 46.6&	16.8&	46.1&	10.8&	36.9 \\
$\texttt{Cap2Det}^{*}$ \cite{Ye:ICCV19}& - &	- &	20.1&	20.3&	20.1 & -&	-&	-&	-&	- \\
\texttt{XPM} \texttt{(Ours)} & $46.8$&	$\textbf{29.9}^{\textbf{\color{blue}+2.4}}$&	$46.3$&	$\textbf{27.0}^{\textbf{\color{blue}+4.2}}$&	$\textbf{41.2}$ & $\textbf{47.3}$&	$\textbf{33.2}^{\textbf{\color{blue}+7.3}}$&	$46.3$&	$\textbf{29.9}^{\textbf{\color{blue}+9.2}}$&	$\textbf{42.0}$ \\
\hline
\end{tabular}
}
}
\label{tab:detection}
\vspace{-0.1cm}
\end{table*}
\begin{table*}[tb]
\centering
\small
\caption{
\footnotesize{
Instance Segmentation (mAP) performances in MS-COCO and Open Images \& Conceptual Captions datasets.
}
}
\vspace{-4pt}
\resizebox{0.95\textwidth}{!}{
\setlength{\tabcolsep}{2pt}{
\begin{tabular}{c|c c|c c c|c c|c c c}
\hline
\multirow{3}{*}{Method} & \multicolumn{5}{c|}{MS-COCO} & \multicolumn{5}{c}{Open Images \& Conceptual Captions}\\
\cline{2-11}
& \multicolumn{2}{c|}{Constrained} & \multicolumn{3}{c|}{Generalized} & \multicolumn{2}{c|}{Constrained} & \multicolumn{3}{c}{Generalized} \\
\cline{2-11}
& \;\; Base \;\; & \;\; Target \;\; & \;\; Base \;\; & \;\; Target \;\; & \;\; All \;\; & \;\; Base \;\; & \;\; Target \;\; & \;\; Base \;\; & \;\; Target \;\; & \;\; All \;\; \\
\hline
\multicolumn{11}{c}{\textit{Caption Pretraining with \cite{Zareian:CVPR21}}} \\
\hline
\texttt{OVR}\cite{Zareian:CVPR21} & 42.0&	20.9&	\textbf{41.6}&	17.1&	35.2&	52.6&	23.8&	45.6&	17.5&	36.2 \\
\texttt{SB}\cite{Bansal:ECCV18} & 41.6&	20.8&	41.0&	16.0&	34.5&	52.8&	24.8&	46.4&	17.3&	36.6 \\
\texttt{BA-RPN}\cite{Zheng:CVPR21} & 41.8&	20.1&	41.3&	15.4&	34.5&	52.9&	25.3&	47.3&	16.9&	37.1 \\
\texttt{OVR+OMP}\cite{Biertimpel:ICCV20} & 31.3&	14.1&	30.5&	8.3&	24.7&	52.5&	24.9&	47.1&	16.8&	36.9 \\
\hline
\multicolumn{11}{c}{\textit{Pseudo-Labeling}} \\
\hline
\texttt{Soft-Teacher}\cite{Xu:ICCV21} & 41.8&	14.8&	41.5&	9.6&	33.2&	52.0&	25.9&	46.6&	17.6&	36.8\\
\texttt{Unbiased-Teacher}\cite{Liu:ICLR21} & 41.8&	15.1&	41.4&	9.8&	33.1&	51.7&	22.2&	45.3&	14.5&	34.9 \\
\texttt{XPM} \texttt{(Ours)} & $\textbf{42.4}$&	$\textbf{24.0}^{\textbf{\color{blue}+3.1}}$&	$41.5$&	$\textbf{21.6}^{\textbf{\color{blue}+4.5}}$&	$\textbf{36.3}$& $\textbf{55.1}$&	$\textbf{31.6}^{\textbf{\color{blue}+5.7}}$&	$\textbf{49.8}$&	$\textbf{22.7}^{\textbf{\color{blue}+5.1}}$&	$\textbf{40.7}$ \\
\hline
\end{tabular}
}
}

\label{tab:instance_seg}
\vspace{-5mm}
\end{table*}

\subsection{Experimental Setup}

\myparagraph{Datasets}.
Following the setup of \cite{Zareian:CVPR21}, we perform experiments on MS-COCO \cite{Lin:ECCV14}, which contains 48 base classes with mask annotations and 17 target classes for evaluation.
The dataset is partitioned into 107,761 training images with 665,387 mask annotations from base classes and 4,836 testing images consisting of 28,538 and 4,614 mask instances for base and target classes, respectively. For captioned images, we use the entire MS-COCO training set with 118,287 images. Each image is annotated with five captions
describing the visually-grounded objects in the image.

To show the effectiveness of our method on large numbers of images and classes, we use large-scale datasets: Open Images \cite{OpenImages:16} with 2.1M instance masks for 300 classes, and Conceptual Captions \cite{Sharma:ACL18} with 3M captioned images. We propose to split Open Images classes into 200 most common classes as base classes with mask annotations while leaving the remaining 100 rarest classes as target classes unknown to the model during training. Thus, we simulate the real-world setting where the rare classes might be unknown during training.

\myparagraph{Evaluation Metrics.}
For both detection and segmentation experiments, we report the mean Average Precision (mAP) at intersection-over-union (IoU) of 0.5 following conventional zero-shot settings \cite{Bansal:ECCV18,Zheng:CVPR21,Zareian:CVPR21}.
To analyze the performances on base and target classes, we measure the mAP scores in two settings: i) \textit{constrained setting} where the model is only evaluated on test images belonging to either base classes or target classes; 
ii) \textit{generalized setting} in which a model is tested jointly on both base and target class images.
The latter setting is more challenging as it requires the model to segment target classes and avoid the base-class bias where the model detects target classes as base classes with high confidence.

\myparagraph{Baselines.}
We compare with \texttt{SB} \cite{Bansal:ECCV18}, which assigns a non-zero background embedding with norm one to predict different background score per bounding box, and open vocabulary object detection \texttt{OVR} \cite{Zareian:CVPR21}, which pretrains its embedding space on caption-image pairs.
To compare with conventional pseudo-labeling baselines, we adapt \texttt{Soft-Teacher} \cite{Xu:ICCV21} and \texttt{Unbiased-Teacher} \cite{Liu:ICLR21}, which only use visual modality to construct pseudo labels, by using embedding heads for novel class recognition.
In addition, we include the state-of-the-art \texttt{BA-RPN} \cite{Zheng:CVPR21} for zero-shot instance segmentation, which proposes to synchronize background classifier between region proposal network and detection heads to reduce background confusion.
We also combine \texttt{OMP} \cite{Biertimpel:ICCV20} with \texttt{OVR}, which augment the class-agnostic mask head with spatial attention features from embedding head.
Finally, to learn from captions images, we compare with \texttt{Cap2Det} \cite{Ye:ICCV19}, which produces pseudo labels for only target and base classes.

\myparagraph{Implementation Details.}
To be comparable with \cite{Zareian:CVPR21}, we use \texttt{Mask R-CNN} architecture with \texttt{ResNet50} backbone from \texttt{maskrcnn-benchmark} code base.
For training the teacher model, we pretrain the backbone following \cite{Zareian:CVPR21} for 150k iterations on MS-COCO and 200k on Conceptual Captions using 8 V-100 GPUs with the batch size of 32 and the initial learning rate of 0.01.
Then we finetune the backbone on segmentation/detection tasks with the batch size of 8 for 90k iterations and the learning rate of 0.001 on both MS-COCO and Open Images datasets to obtain the teacher model. The student is initialized with teacher weights and trained on pseudo and ground-truth labels for an additional 70k iterations.
We also downweight the detection loss for background class to 0.2 to improve the recall of target classes, similar to \cite{Zareian:CVPR21}.
For the robust student model, we set $\eta = 0.01$, which is the smallest average noise level estimated offline on 10k captioned images.
We use the word embeddings from \texttt{BERT} trained on BookCorpus, and English Wikipedia \cite{Devlin:NAACL19}.
For training the noise estimation module, $g_{\text{Noise}}$, we use reparametrization trick \cite{Kingma:ICLR14} to backpropagate gradients through the sampled noise value, $\epsilon$.
Moreover, we do not optimize $g_{\text{noise}}$ with respect to $\mL^{\alpha}_{X}$, which would result in the trivial solution where the student always predicts low-reliability scores to minimize the loss.

\subsection{Experimental Results}

\myparagraph{Object Detection.}
We evaluate our method for the object detection task under bounding box or mask supervision of base classes in Table \ref{tab:detection} on MS-COCO. Based on the base/target class results in the constrained setting and the generalized setting, we make the following conclusions:

\point Although using caption pretraining improves the performance on target classes (under bounding box supervision) over zero-shot training, this strategy does not work as well for mask-level supervision. Since caption-based backbone pretraining \cite{Zareian:CVPR21} can only learn high-level spatially-coarse features of objects but not fine-grained object masks, finetuning on mask annotations corrupts the learned backbone and degrades its performances on target classes. This shows the incompatibility between the mask prediction task and the information encoded in the pretrained backbone.

\point \texttt{Soft-Teacher} and \texttt{Unbiased Teacher} improve the performance on base classes (under box-level supervision) over using caption pretraining alone.
However, as these baselines do not constrain their predictions based on captions, they miss-label novel classes, which propagates teacher error and degrades target-class performances. Although \texttt{Cap2Det} conditions its pseudo labels on captions, these labels come from a limited set of base and target classes. Thus, \texttt{Cap2Det} cannot exploit the useful information from other novel objects in the captions.

\point With bounding box supervision, our method (without estimating mask noises) significantly improves target class performances by 2.4\% and 4.2\% in constrained and generalized settings, respectively.
This shows the importance of leveraging captions to improve the pseudo labeling of target classes without annotations.
Moreover, with additional mask annotations, we further gain 9.2\% in performance on target classes compared to state-of-the-art, which shows the effectiveness of self-training on pseudo masks.

\begin{figure*}[t!]
\centering
\includegraphics[width=0.9\textwidth]{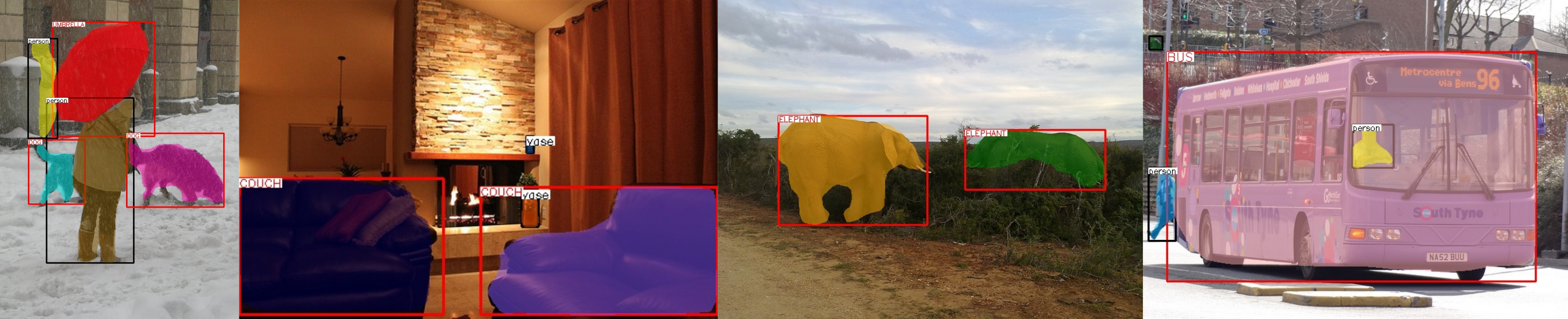}
\vspace{-2mm}
\caption{\footnotesize{
Visualization of our mask predictions for base classes (in \textbf{back box}) and target classes (in \textbf{\color{red} red box}) in the generalized setting.
}}
\label{fig:qualitative}
\vspace{-4mm}
\end{figure*}

\begin{figure*}[t!]
\centering
\includegraphics[width=0.9\textwidth]{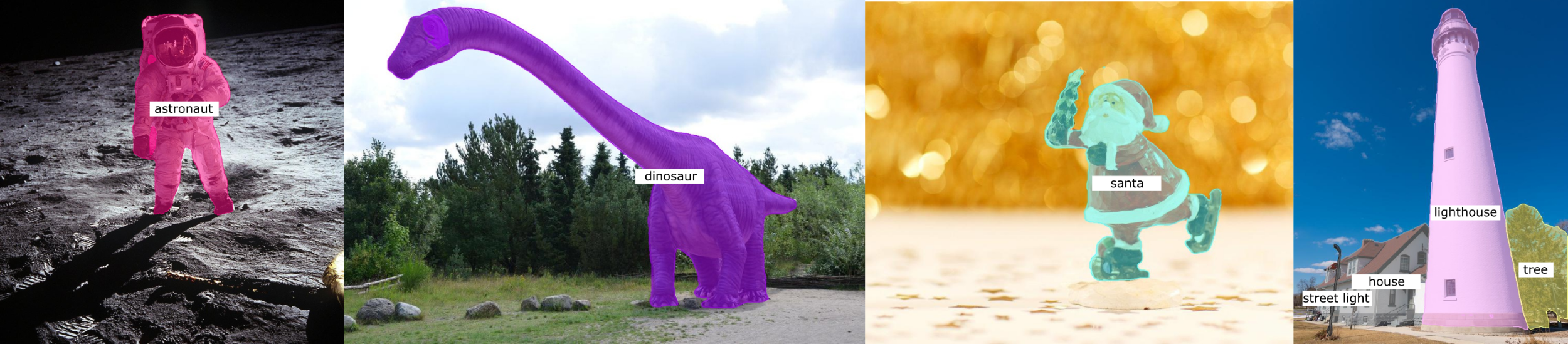}
\vspace{-2mm}
\caption{\footnotesize{
Visualization of our mask predictions for novel classes in the wild with large-scale cross-modal pseudo-labeling.
}}
\label{fig:qualitative_large_scale}
\vspace{-4.5mm}
\end{figure*}
\begin{figure}
\begin{minipage}[b]{0.50\linewidth}
\includegraphics[height=25mm]{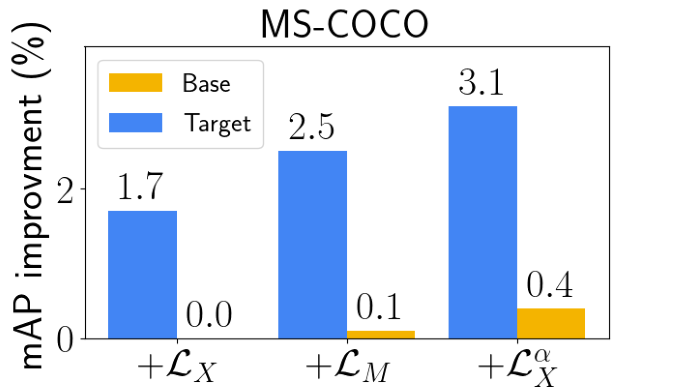}
\end{minipage}
\begin{minipage}[b]{0.42\linewidth}
\includegraphics[height=25mm]{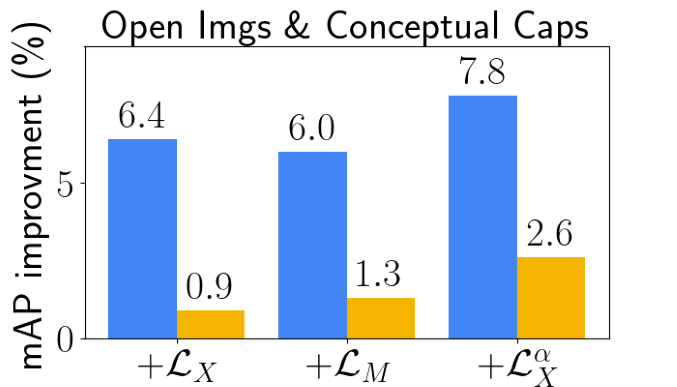}
\end{minipage}
\vspace{-2.0mm}
\caption{
\footnotesize{Segmentation improvements w.r.t. the teacher model from adding different proposed components to the student.
}
}
\label{fig:ablation}
\vspace{-2.0mm}
\end{figure}

\myparagraph{Instance Segmentation.}
To show the effectiveness of \texttt{XPM}, we conduct instances segmentation experiments on both MS-COCO and Open Images datasets. We report the results in Table \ref{tab:instance_seg} and conclude that:

\point On MS-COCO, different background modeling techniques in \texttt{SB, BA-RPN} have minimal impact on target-class performances when combined with embedding-based caption pretraining.
On the other hand, explicitly transferring this knowledge from embedding to mask heads via \texttt{OMP} significantly degrades the performances on base and target classes. This is due to the insufficient amount of base classes and training samples to learn meaningful Object Mask Prior from the small-scale MS-COCO dataset.

\point On the large-scale Conceptual Captions and Open Images datasets, both \texttt{SB} and \texttt{BA-RPN} improve target-class segmentation in the constrained setting, as more accurate background models can be learned from the larger number of base classes in Open Images compared to MS-COCO.
We observe that conventional pseudo-labeling methods \texttt{Soft-Teacher, Unbiased Teacher} have no significant improvements over caption-pretraining baselines since they cannot utilize textual modality in captioned images to spot novel classes correctly.

\point Overall, our method achieves significant performance improvements of at least 4.5\% and 5.1\% mAP score compared to other baselines in MS-COCO and Open Images datasets, respectively.
Moreover, in the Conceptual Captions and Open Images setting, we observe a compound effect -- as a result of using a larger number of base classes for training, our teacher model generalizes significantly better on target classes.
When labeling Conceptual Captions with the teacher, we benefit from the significant increase in pseudo labels' quantity and quality.
Thus, the student obtains strong results on both base and target classes, with a significant gain of 3.6\% on all classes.

\myparagraph{Ablation Study.}
Figure \ref{fig:ablation} shows our segmentation improvements compared to the teacher model when introducing different components in our method, on both MS-COCO and Open Images \& Conceptual Captions.
Adding the cross-modal loss, $\mL_{X}$, significantly improves the segmentation performance over the teacher model, as the student can distill rich knowledge from captioned images.
Although the mask loss, $\mL_{M}$, improves target-class performance on MS-COCO, it fails to improve with Conceptual Captions due to noisy web captions.
By regulating the cross-modal loss with noise estimation, $\mL_{X}^\alpha$, we gain further improvement on both caption datasets by mitigating the error propagation from teacher to student model.

\myparagraph{Effectiveness of Robust Student.}
In Table \ref{tab:uncertainty_measurement}, we experiment with other methods on pseudo-mask noise estimation and loss reweighting. We evaluate \texttt{Stochastic BCE} \cite{Hu:NeurIPS20} which learns pixel-wise noise to regulate mask loss, $\mL_{M}$.
This method is unable to improve performances as it cannot use mask noise to regulate the cross-modal loss for classification.
For the methods that regulate cross-modal loss $\mL_{X}^\alpha$, we consider \texttt{Class Score} which uses class prediction confidences, \texttt{Pixel Score} \cite{Yang:Arxiv21} which estimates mask quality by aggregating pixel-wise prediction confidence, and \texttt{DropOut Entropy} \cite{Gal:ICML16} which computes prediction entropy via multiple dropout passes.
These methods provide no significant improvements, as they are trained on clean annotations of base classes and not adapted to noisy pseudo masks.
By learning to estimate the noise levels of pseudo masks and regulating both $\mL_{X},\mL_{M}^\alpha$, we achieve superior performances compared to \texttt{No Noise Estimation}.

\begin{table}
\small
\centering
\caption{\footnotesize{Segmentation performances of different strategies for noise estimation and loss weighting on Open Images.
}
}
\vspace{-4pt}

\setlength{\tabcolsep}{1.5pt}{
\begin{tabular}{l|c|ccc}
\hline
\multirow{1}{*}{Method} & Used on & Base & Target & All\\
\hline\hline
\multirow{1}{*}{\texttt{No Noise Estimation}} & - & 53.3 &	30.2& 39.1 \\
\hline
\multirow{1}{*}{\texttt{Stochastic BCE} \cite{Hu:NeurIPS20}}& $\mL_{M}$ &	53.8& 29.8&	39.2\\
\hline
\multirow{1}{*}{\texttt{Class Score}}&\multirow{3}{*}{$\mL_{X}^\alpha$} & 54.0 & 28.4 &	38.8\\
\multirow{1}{*}{\texttt{Pixel Score} \cite{Yang:Arxiv21}} &&53.2&	30.1&	38.5 \\
\multirow{1}{*}{\texttt{DropOut Entropy} \cite{Gal:ICML16}}&&	53.6&	29.7&	38.5\\
\hline
\makecell{\texttt{Robust Student (Ours)}} & $\mL_{X}^\alpha+\mL_{M}$ & \textbf{55.1} & \textbf{31.6} &	\textbf{40.7}\\
\hline
\end{tabular}
}
\label{tab:uncertainty_measurement}
\vspace{-4.5mm}
\end{table}
%

%

\myparagraph{Qualitative Results.} Figure \ref{fig:qualitative} shows the mask predictions of our methods for both base and target classes on MS-COCO.
Our method can correctly detect and segment multiple instances of target classes without any ground-truth mask annotations during training.
Moreover, our framework maintains strong performances on base classes such that it can correctly segment the base class ``bus driver'' (the last example) within the target class ``bus''.

We also visualize the pixel-wise noises for each object in captions in Figure \ref{fig:mask_noise}.
We observe that a good pseudo mask (\textit{e.g.,} `bear') only has a few noisy pixels along its object boundaries. Whereas, an incorrect pseudo mask (\textit{e.g.,} `skateboard') contains a large number of noisy pixels that spread over large areas within the bounding boxes.

\myparagraph{Large-Scale Cross-Modal Pseudo-Labeling.}
To demonstrate the scalability of our method, we apply cross-modal pseudo-labeling with multiple segmentation datasets (Open Images\cite{OpenImages:16}, LVIS\cite{Gupta:CVPR19}), object detection dataset (Objects365\cite{Shao:ICCV19}), and caption dataset (Conceptual Captions\cite{Sharma:ACL18}), to create a high-performance student model.
As shown in Figure \ref{fig:qualitative_large_scale}, this strong student, trained with our method, successfully generalized to novel classes such as ``astronaut'' and ``dinosaur'', which never appear in most segmentation datasets.
Moreover, we can segment the fine details of such truly novel classes without any mask annotations.

\section{Conclusions}
We tackle the problem of open-vocabulary instance segmentation by proposing a robust cross-modal pseudo-labeling framework to provide mask supervision of novel classes in captioned images for training segmentation models.
We show the effectiveness of our method on both MS-COCO and Open Images \& Conceptual Captions datasets.
However, our method might not be suitable for learning with limited base classes as we assume the base classes are sufficiently diverse to enable novel-class generalization.

\section*{Acknowledgement} 
\vspace{-2mm}
We would like to thank Ping Hu for his valuable suggestions on implementing the robust student model.
This work is partially supported by DARPA (HR00112220001), NSF (IIS-2115110) and ARO (W911NF2110276). Content does not necessarily reflect the position/policy of the Government. No official endorsement should be inferred.

\newpage
{\small
\bibliographystyle{IEEEtran}
\bibliography{./biblio_bank/ehsan,./biblio_bank/multilabellearning,./biblio_bank/zeroshot_learning,./biblio_bank/fewshot_learning,./biblio_bank/recognition,./biblio_bank/vision,./biblio_bank/generative_model,./biblio_bank/learning,./biblio_bank/neuralnet,./biblio_bank/nlp,./biblio_bank/attention,./biblio_bank/hoi,./biblio_bank/composition,./biblio_bank/detection,./biblio_bank/robustness,./biblio_bank/vision_language}
}
\end{document}